\documentclass[runningheads]{llncs}
\usepackage{graphicx,amsmath,multirow}
\usepackage{amsfonts}
\usepackage{amssymb}
\usepackage{subfigure}
\usepackage{footnote}
\usepackage{enumerate}
\usepackage{longtable}
\usepackage{multirow}
\usepackage{array}
\newcolumntype{C}[1]{>{\centering\let\newline\\\arraybackslash\hspace{0pt}}m{#1}}
\usepackage{booktabs}
\usepackage{hyperref}
\usepackage[figuresright]{rotating}
\usepackage[ruled]{algorithm2e}
\usepackage{color}

\usepackage{graphicx}
\begin{document}
\title{Inductive Semi-supervised Learning Through Optimal Transport}
%
%
\author{Mourad El Hamri\inst{1,2} \and
Younès Bennani\inst{1,2} \and
Issam Falih\inst{2,3}}
\authorrunning{El Hamri et al.}
%
\institute{LIPN - CNRS UMR 7030, Université Sorbonne Paris Nord, France \and
LaMSN - La Maison des Sciences Numériques, France\and
LIMOS - CNRS UMR 6158,  Université Clermont Auvergne, France
\email{\{Firstname.Lastname\}@\{sorbonne-paris-nord,uca\}.fr}}
\maketitle              
\begin{abstract}
In this paper, we tackle the inductive semi-supervised learning problem that aims to obtain label predictions for out-of-sample data. The proposed approach, called Optimal Transport Induction (OTI),  extends efficiently an optimal transport based transductive algorithm (OTP) to inductive tasks for both binary and multi-class settings. A series of experiments are conducted on several datasets in order to compare the proposed approach with state-of-the-art methods. Experiments demonstrate the effectiveness of our approach. We make our code publicly available.\footnote{Code is available at: https://github.com/MouradElHamri/OTI}
\end{abstract}
\section{Introduction}
Supervised learning models, especially deep neural networks can 
achieve human or superhuman level performances on a broad spectrum of learning tasks. This big achievement, however, heavily relies on the availability of large-scale labeled datasets, which usually come at a significant cost, since labeling data often requires an extensive human labor. For instance, the work needed for labeling manually sequential data (e.g. speech or video) is proportional to their lengths. Moreover, some specific domain knowledge (e.g. medicine) is often critical for labeling and requires an involvement of experts. This lack of labeled data is often accompanied by an abundance of unlabeled instances, which raises the question of how to take advantage of the big amount of unlabeled samples to alleviate the need for large high quality labeled datasets. 
\\
\\Semi-supervised learning (SSL) \cite{chapelle2009semi}  has emerged as one of the most promising paradigms to achieve this goal by leveraging both labeled and unlabeled data. There are two categories of semi-supervised learning models \cite{van2020survey}: inductive learning models and transductive learning models. Inductive semi-supervised learning aims to learn  using both labeled and unlabeled data a predictive model capable of making predictions both for unlabeled data already encountered in the training phase and for out-of-sample data, i.e. previously unseen instances. While transductive semi-supervised learning methods are only concerned with predicted the labels of unlabeled data already presented during the training. 
\\
\\Most of transductive learning models are graph-based approaches \cite{subramanya2014graph} (e.g. label propagation methods), that aim to present each data, labeled or not by a vertex in the graph, then, similarities between vertices are evaluated to determine whether or not there should be an edge between each pair of vertices, finally, edges are weighted to reflect the similarity degree between vertices. After the graph construction, labels are inferred in several ways (e.g. in the case of label propagation, labels are diffused from labeled to unlabeled data through the graph edges). The common major inconvenience of transductive methods is their inability to predict labels for out-of-sample data, so when some previously unseen test data arrive, transductive learning methods need to fusion these new samples into the previous data in our disposal to reconstruct a new augmented graph based on merged data, and then perform label propagation from scratch. This process is too costly, since the presentation of even, a single new point require to rerun all the process already done in their entirety, which is distasteful in many real-world applications, where on-the-fly prediction for previously unseen instances is highly requested. 
\\
\\ Several works have attempted to bridge the gap between transductive and inductive semi-supervised learning, by proposing to derive an inductive function for out-of-sample data from the transductive predictions \cite{delalleau2005efficient} \cite{bengio2006}. However, these approaches  cannot be generalized to all the transductive methods, such as, Optimal Transport Propagation (OTP) \cite{el2021label}. This inability to generalize is mainly due to the difference of this approach from the state-of-the-art methods, since instead of a complete graph, OTP relies on a bipartite graph, which modifies its objective function, that does not follow the standard form of objective functions composed usually of two terms, a first term to penalize predicted labels that do not match the true labels and another term to penalize the difference in labels between similar samples. Such a setting motivated the development of a simple way to extend OTP (and other transductive semi-supervised approaches based on bipartite graphs) to the inductive framework. 
\\ 
\\ In this paper, we derive an inductive approach called Optimal Transport Induction (OTI) from the modified objective function of OTP.  In addition to being applicable in the case of binary classification, OTI can be efficiently extended to multi-class settings. 
\\
\\The rest of this paper is organized as follows: in Section 2, we present an overview of optimal transport \cite{villani2008optimal} \cite{santambrogio2015optimal}. Section 3 details the proposed optimal transport induction method (OTI) and in Section 4, we provide comparisons to state-of-the-art methods on several benchmark datasets.
\section{Optimal Transport}
The birth of optimal transport is dated back to 1781, with the following problem introduced by Gaspard Monge \cite{monge1781memoire}: Let $(\mathcal{X},\mu)$ and $(\mathcal{Y},\nu)$  be two probability spaces and $c : \mathcal{X}\times\mathcal{Y} \to \mathbb{R}^+ $ a measurable cost function, the problem of Monge aims at finding the transport map $\mathcal{T} : \mathcal{X} \to \mathcal{Y}$, that transport the mass represented by the measure $\mu$ to the mass represented by the measure $\nu$ and which minimizes the total cost of this transportation, more formally:
\begin{equation}  \quad \underset{\mathcal{T}}{\inf}\{\int_{\mathcal{X}}  c(x,\mathcal{T}(x)) d\mu(x) |  \mathcal{T}\#\mu = \nu \},   \end{equation}
where $\mathcal{T}\#\mu$ denotes the push-forward operator of $\mu$ through the map $\mathcal{T}$. 
\\
\\A long period of sleep followed Monge's formulation until the relaxation of Leonid Kantorovitch in 1942 \cite{kantorovich1942translocation}. The relaxed formulation of kantorovich, known as the Monge-Kantorovich problem, can be formulated in the following way:
\begin{equation} \underset{\gamma}{\inf} \{\, \int_{\mathcal{X}\times\mathcal{Y}} \, c(x,y) \, d\gamma(x,y) \,|\, \gamma \in \Pi(\mu,\nu)\, \}, \end{equation}
where $\Pi(\mu,\nu)$ is the set of probability measures over the product space $\mathcal{X}\times\mathcal{Y}$ such that both marginals of $\gamma$ are $\mu$ and $\nu$.
\\
\\ In several real world application, the access to the measures $\mu$ and $\nu$ is only available through finite samples $X = (x_1,...,x_n) \subset \mathcal{X}$ and $Y = (y_1,...,y_m) \subset \mathcal{Y}$, then, the measures $\mu$ and $\nu$ can be casted as the following discrete measures, $\mu = \sum_{i=1}^n a_{i} \delta_{x_{i}}$ and
$\nu = \sum_{j=1}^m b_{j} \delta_{y_{j}}$, where $a \in \sum_n$ and $b \in \sum_m$ are probability vectors of size $n$ and $m$ respectively. The relaxation of Kantorovich becomes then the following linear program \cite{peyre2019computational}: 
\begin{equation}
\underset{T \in U(a,b)}{\min}  \langle {T},{C_{XY}} \rangle _F 
\end{equation}
where  $U(a,b) = \{T \in \mathcal{M}_{n \times m}(\mathbb{R}^{+}) \, | \, T 1_{m} = a \,\, \text{and} \,\, T^{\mathbf{T}} 1_{n} = b\}$ is the transportation polytope which acts as a feasible set, $C_{XY}$ is the cost matrix and $ \langle T,C_{XY} \rangle _F \,= trace(T^\mathrm{T}C_{XY})$ is the Frobenius dot-product of matrices. 
\\
\\ This linear program, can be solved with the simplex algorithm or interior point methods. However, optimal transport problem scales cubically on the sample size, which is often too costly in practice, especially for machine learning applications that involve massive datasets. Entropy-regularization \cite{cuturi2013sinkhorn} has emerged as a solution to the computational burden of optimal transport. The entropy-regularized discrete optimal transport problem reads:
\begin{equation}
    \underset{T \in U(a,b)}{\min}  \langle {T},{C_{XY}} \rangle _F - \varepsilon \mathcal{H}(T)   
\end{equation}
where  $\mathcal{H}(T) = - \sum_{i=1}^n \sum_{j=1}^m t_{ij} (\log(t_{ij}) - 1)  $ is the entropy of $T$. This regularized problem can be solved efficiently via an iterative procedure: Sinkhorn-Knopp algorithm.
\section{Optimal Transport Induction}
In this section, we propose an efficient way to extend OTP \footnote{which would be too lengthy to detail here, please refer directly to this approach in the following paper \cite{el2021label}.} for out-of-sample data. The main underlying idea behind this extension is the modification of the objective function of OTP, from which we derive a novel algorithm called Optimal Transport Induction (OTI) able to predict labels for previously unseen data.
\newline
\newline The transductive method OTP can be casted as the minimization of the following objective function $\mathcal{C}^{transduction}_{\mathcal{W},l_u}$ in terms of the label function values at unlabeled samples $x_j \in X_U$:
\begin{equation}
   \mathcal{C}^{transduction}_{\mathcal{W},l_u}(f) = \sum_{x_i \in X_L} \sum_{x_j \in X_U} w_{x_i,x_j} l_u(y_i,f(x_j))
\end{equation}
where $l_u$ is an unsupervised loss function (e.g. a lower-bounded dissimilarity function applied on a
pair of output values $y_i$ and $f(x_j)$), and $\mathcal{W} = (w_{x_i,x_j})_{i,j}$ is the affinity matrix derived from the optimal transport plan between labeled and unlabeled data. The objective function in Eq(5) is a smoothness criterion that seeks to penalize differences in the label predictions for similar data points in the graph, which means that a good classifier should not change too much between similar samples. The main difference between the objective function of OTP and traditional transductive approaches is the absence of a second term to penalize labels that do not match the correct ones, because OTP relies on a bipartite edge weighted graph instead of a fully connected graph to leave the labels $Y_L$ unchanged during the training. The absence of this term in the objective function of OTP, makes the use of state-of-the-art induction formulas unfeasible. which is the main motivation behind the conception of OTI. 
\subsection{Problem setup}
Let $X=\{x_1,...,x_{l+u}\}$ be a set of $l+u$ data points in $\mathbb{R}^d$ and $\mathcal{C} =\{c_1,...,c_K\}$ a discrete label set consisting of $K$ classes. The first $l$ points denoted by $X_L=\{x_1,...,x_l\}$ are labeled according to $Y_L=\{y_1,...,y_l\}$, where $y_i \in \mathcal{C}$ for every $i \in \{1,...,l\}$, and the remaining data points denoted by $X_U=\{x_{l+1},...,x_{l+u}\}$ are labeled using OTP according to  $Y_U=\{y_{l+1},...,y_{l+u}\}$. For inductive classification, unseen data in the training phase are denoted by $X_{new}$. The aim of OTI is to predict labels of $X_{new}$ using $X=X_L \cup X_U$ and $Y=Y_L \cup Y_U$, without being obliged to perform OTP from scratch.
\subsection{Induction formula: OTI}
In order to extend the transductive algorithm OTP into function induction for out-of-sample data points, the same smoothness criterion as in Eq(5) will be used for new test instances $x_{new} \in X_{new}$ \cite{bengio2006}, and then we can minimize the modified objective function with respect to only the predicted labels $\Tilde{f}(x_{new})$. The novel smoothness criterion for  new test samples $x_{new}$ becomes then:
\begin{equation}
\mathcal{C}^{induction}_{\mathcal{W},l_u}(\Tilde{f}(x_{new}))  =  \sum_{x_i \in X_L \cup X_U} w_{x_i,x_{new}} l_u(y_i,\Tilde{f}(x_{new}))
\end{equation}
If the loss function $l_u$ is convex, e.g., $l_u = (y_i -\Tilde{f}(x_{new}))^2$, then the cost function $\mathcal{C}^{induction}_{\mathcal{W},l_u}$ is also convex in $\Tilde{f}(x_{new})$. Thus the label assignment $\Tilde{f}(x_{new})$ minimizing $\mathcal{C}^{induction}_{\mathcal{W},l_u}$ is given by:
\begin{equation}
\Tilde{f}(x_{new}) = \frac{\sum_{x_i \in X_L \cup X_U} w_{x_i,x_{new}} y_i}{\sum_{x_i \in X_L \cup X_U} w_{x_i,x_{new}}}  
\end{equation}
The similarities $w_{x_i,x_{new}}$ are directly inferred using optimal transport as follows:
\begin{equation}
w_{x_i,x_{new}} = \frac{\gamma^*_{\varepsilon_{x_i,x_{new}}}}{\sum_i \gamma^*_{\varepsilon_{x_i,x_{new}}}}
\, \, \, \forall x_i\in X=X_L \cup X_U,
\end{equation}
where $\gamma^*_\varepsilon$ is the new optimal transport plan between the empirical distributions of $X=X_L \cup X_U$ and $X_{new}$ defined by:
\begin{center}
    $\gamma^*_\varepsilon = \underset{\gamma \in U(a,b)}{argmin} \,\, \langle {\gamma},{C} \rangle _F - \varepsilon \mathcal{H}(\gamma),$
\end{center}
and $C$ is the cost matrix between $X=X_L \cup X_U$ and $X_{new}$:
\begin{center}
    $C=[c_{x_i,x_{new}}] \, \text{defined by} \,\, c_{x_i,x_{new}}= \lVert x_i - x_{new} \rVert^2, \, \forall (x_i,x_{new}) \in X \times X_{new}$
\end{center}
\subsection{Binary classification and multi-class settings}
In binary classification, where $\mathcal{C} =\{+1,-1\}$, the classification problem in Eq(7) can be casted as a regression problem, in the following way:
\begin{equation} 
\begin{cases} 
y_{new}  = +1  \,\,\, \text{if} \,\, sign(\Tilde{f}(x_{new})) \ge 0\\
y_{new}  = -1  \,\,\, \text{otherwise} 
\end{cases},  
\end{equation}
Most of transductive approaches possess the ability to handle multiple classes, while the inductive methods are usually limited to the binary classification framework, where $\mathcal{C} =\{+1,-1\}$ as in Eq(9). However, our proposed approach OTI can be adapted accurately for multi-class settings, as follows: the label $\Tilde{f}(x_{new})$ is given by a weighted majority vote of the training samples in $X = X_L \cup X_U$:

\begin{equation}
\Tilde{f}(x_{new}) = \underset{c_{k} \in \mathcal{C}}{argmax}  \sum_{x_i \in X_L \cup X_U / y_{i} = c_{k}} w_{x_i,x_{new}}
\end{equation}
The predicted class of $x_{new}$ is then the class whose representatives have the highest similarity with $x_{new}$.
\\
\\ Eq(9) can be seen as a special case of Eq(10) in the binary classification settings, in fact, if $\mathcal{C} =\{+1,-1\}$, then choosing between the class that maximizes $\sum_{x_i \in X_L \cup X_U / y_{i} = c_{k}} w_{x_i,x_{new}}$ is equivalent to choosing according to the sign of $\Tilde{f}(x_{new})$, since the term $\sum_{x_i \in X_L \cup X_U } w_{x_i,x_{new}}$ in $\Tilde{f}(x_{new})$ is always positive. 
\\
\\ The proposed algorithm OTI, is formally summarized in algorithm 1, where we use the algorithm OTP for training and the induction formula for testing.
\vspace{-6mm}
\begin{algorithm}\label{OT-ISSL}
\caption{OTI}
\SetKwInOut{Input}{Input}
\SetKwInOut{Parameters}{Parameters}
\SetKwInOut{Output}{Output}
\Input{$X_{new}, X_L, X_U, Y_L$}
\Parameters{$\varepsilon$}
\textbf{(1) Training phase}  \\
Compute $Y_U$ by OTP \\
\textbf{(2) Testing phase}  \\
\For{\text{a new test point} $x_{new} \in X_{new} $}{
Compute $w_{x_i,x_{new}} \,\,\, \forall x_i \in X = X_L \cup X_U$ by Eq(8) \\
Compute the label $\Tilde{f}(x_{new})$ by Eq(9) or Eq(10)  \\
}
\Return{$\Tilde{f}(x_{new})$}
\end{algorithm}
\vspace{-12mm}
\section{Experiments}
 \subsection{Experimental protocol}
In this section, we provide empirical experimentation for the proposed algorithm OTI.
To thoroughly evaluate the performance of the proposed approach, a total of five benchmark datasets \textbf{(Iris, Ionosphere, Dermatology, Digits, MNIST)} have been employed for experimental studies \footnote{The datasets are publicly available at: https://archive.ics.uci.edu/ml/datasets.php}. 
The performance of \textbf{OTI} is compared with four state-of-the-art methods, including three transductive and one inductive semi-supervised learning approaches: \textbf{LP} \cite{zhu2002learning}, \textbf{LNP} \cite{wang2007label}, \textbf{OTP} \cite{el2021label}  and \textbf{SSI} \cite{delalleau2005efficient}.
To evaluate the performance of our approach, two widely-used evaluation measures were employed: the normalized mutual information \textbf{(NMI)} \cite{dom2012information}, and the adjusted rand index \textbf{(ARI)} \cite{hubert1985comparing}. 
For each dataset, we randomly sample $\zeta\times100\%$ instances to form the labeled set $X_L$. For the remaining examples, $40\%$ of them are randomly sampled to form
the unlabeled set $X_U$ and $60\%$ of them are randomly sampled to
form the out-of-sample data $X_{new}$. In the experiments, the training and testing procedures are conducted for the compared algorithms as follows: for the inductive semi-supervised learning methods OTI and SSI, both of them are trained on $X = X_L \cup X_U$ and tested on $X_{new}$. The three transductive semi-supervised learning methods LP, LNP and OTP can only make predictions on unlabeled samples already encountered in the training, so, to ensure that all the comparing algorithms are evaluated on the same out-of-sample set, the three algorithms are trained on $\tilde{X} = X_L \cup X_U \cup X_{new}$ and the evaluation is restricted to $X_{new}$. The sampling rate $\zeta$ for labeled data is varied from 5\% to 25\% with a step-size of 10\%. Under each sampling rate, all the four compared algorithms were run with ten different random sampling, the mean performance out of the ten runs is recorded for the four algorithms. 
\subsection{Results}
Table 1 reports the detailed experimental results of each algorithm on all the datasets in terms of ARI and NMI.

\vspace{-4mm}

\begin{table}[h!]
\caption{Inductive performances in terms of ARI and NMI}
\small
\centering
\setlength\tabcolsep{3pt}
\setlength\extrarowheight{0pt}
\begin{tabular}{|l|c|c|c|c|c|c|c|c|c|c|c|}
\cline{3-12}
\multicolumn{2}{c|}{}&\multicolumn{5}{c|}{ARI} &\multicolumn{5}{c|}{NMI} \\ 
\hline
Datasets    & $\zeta$  & LP    & LNP      & OTP     & SSI     & OTI   & LP    & LNP      & OTP     & SSI     & OTI          \\ \hline 

             & 5   & 0.767  &0.728  &0.772 & 0.720& 0.753 & 0.743  & 0.682  &0.749  & 0.691 & 0.723\\ 
Iris         & 15  & 0.860  &0.804  &0.889 & 0.796 &0.862& 0.863 & 0.774  &0.863 &  0.772& 0.818\\ 
             & 25  & 0.880  &0.857  &0.917 & 0.845 & 0.897& 0.874& 0.849  &0.882  & 0.847 & 0.869\\
\hline             
             & 5   & 0.328  &0.307  &0.527 & 0.298 &0.502 & 0.293 & 0.262 &0.383 & 0.263 & 0.326\\
 Ionosphere  & 15  & 0.449 &0.419  &0.595 &  0.408& 0.572 & 0.373 & 0.347 &0.483&  0.342& 0.424\\
             & 25  & 0.483 & 0.462  &0.618  & 0.460&0.591 & 0.404 & 0.375 &0.530 & 0.362 & 0.445\\ 
\hline     
             & 5  & 0.806 & 0.799 &0.850 & 0.789& 0.837 & 0.814  & 0.795 &0.835  & 0.783 & 0.819\\
 Dermatology & 15  & 0.902 & 0.869 &0.922 & 0.862& 0.901  & 0.893  & 0.854 &0.914 & 0.853  & 0.889  \\
  & 25  & 0.918 & 0.894 &0.945 &0.892 &0.919 & 0.918 &0.882 &0.921 & 0.869  & 0.908 \\                      
\hline 
             
             & 5  & 0.842  &0.820 &0.887 & 0.827 & 0.868 & 0.851 & 0.803 &0.872  & 0.809 & 0.853 \\
Digits               & 15 & 0.931 &0.915 &0.952 & 0.925 &0.930 & 0.937  & 0.901 &0.950 & 0.892 & 0.930 \\
     & 25 & 0.969 &0.958 & 0.980 & 0.960 & 0.968 & 0.962  & 0.949 &0.968 & 0.928 & 0.947\\ 
\hline 
             & 5  & 0.774  &0.737 &0.795 & 0.730 &0.772 & 0.761 & 0.726 &0.777 &  0.719& 0.759 \\
     MNIST   & 15 & 0.811 &0.784 &0.855 & 0.783& 0.836& 0.826 & 0.793  &0.830 & 0.782 &  0.812\\
      & 25&  0.863 & 0.839 &0.884 &0.841 & 0.859& 0.850 & 0.814  &0.861 & 0.810  & 0.849 \\
\hline 
  \multicolumn{2}{|c|}{Average} & 0.772  & 0.746 & 0.826 &0.748 &0.804 & 0.758  & 0.720 & 0.788 &0.715 &0.758 \\  \hline 
\end{tabular}
\label{ARI}
\vspace{-2mm}
\end{table} 
\vspace{-3mm}
\noindent The first remark that can be made from this table is that the performance of each algorithm on all the datasets grows in parallel with the growth of the sampling rate $\zeta$, which is quite normal, because the amount of the available labeled data is an important factor to improve the performance of semi-supervised models, the more labeled data is available, the better the model is performing. The table also shows that our method outperforms the other inductive model SSI, in all the datasets and for all the sampling rates $\zeta$, this achievement can be explained by the same reason why OTP outperforms LP, i.e. the use of optimal transport to calculate similarities between the data instead of the pairwise distances which only capture information on a bilateral level. However, the performance of OTI is inferior to that of OTP, which is reasonable because, on the one hand, in order to allow to OTP to perform inductive tasks, the out-of-sample set $X_{new}$ was merged with the sets $X_L$ and $X_U$ in the training, while only predictions on $X_{new}$ were retained to evaluate its performance, and on the other hand, OTI relies on OTP in the training, so if OTP makes errors in its predictions on $Y_U$, this will directly affect the performance of OTI on $X_{new}$. The same analysis can be extended to the case of SSI, that uses LP in its training, which explains the better performance of the latter. The difference in the average performances for both NMI and ARI, between OTP and OTI, is less than the difference between LP and SSI, demonstrating that using optimal transport to extend the label function to out-of-sample data can significantly improve the accuracy of the predictions.
\section{Conclusion}
In this paper, the problem of inductive semi-supervised learning is addressed. A new approach, named Optimal Transport Induction, is proposed to extend OTP from transductive parameters to inductive ones.  A modification in the objective function of OTP was considered, from which an efficient algorithm OTI was derived to allow us to extend predictions to previously unseen data, by relying on optimal transport to compute the similarity between the out-of-sample data and the data already encountered in the training phase.
Experimental studies have been conducted to show the effectiveness of our approach.
\bibliographystyle{plain}
\bibliography{ICONIP2021.bib}
\end{document}